\documentclass{article}
\usepackage{arxiv}
\usepackage{appendix}
\usepackage{cite}
\usepackage{amsmath,amssymb,amsfonts}
\usepackage{algorithmic}
\usepackage{graphicx}
\usepackage{textcomp}
\usepackage{xcolor}
\usepackage{pgfplots}
\usepackage{xcolor}
\usepackage{listings}
\usepackage{pdfpages}
\usepackage{xcolor}
\usepackage{hyperref}
\usepackage{float}
\usepackage{graphicx}
\usepackage{subcaption}
\usepackage{booktabs}
\usepackage{multirow}
\usepackage{wrapfig}
\usepackage[warning]{numprint}

\usepackage{xspace}
\usepackage{moresize}

\usepackage[utf8]{inputenc} %
\usepackage[T1]{fontenc}    %
\usepackage{hyperref}       %
\usepackage{url}            %
\usepackage{booktabs}       %
\usepackage{amsfonts}       %
\usepackage{nicefrac}       %
\usepackage{microtype}      %
\usepackage{cleveref}       %
\usepackage{lipsum}         %
\usepackage{graphicx}
\usepackage{doi}

\title{Neural Stochastic Differential Equations for Robust and Explainable Analysis of Electromagnetic Unintended Radiated Emissions\\
}

\date{}

\newif\ifuniqueAffiliation
\uniqueAffiliationtrue

\ifuniqueAffiliation %
\author{ 
    {\href{https://orcid.org/0000-0003-0354-2940}{\includegraphics[scale=0.06]{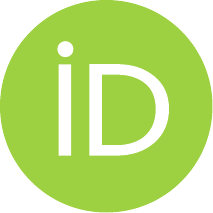}}\hspace{1mm}Sumit Kumar Jha}\\
    Computer Science Department\\
    Florida International University\\
    \texttt{sumit.jha@fiu.edu} \\
    \And
    {\href{https://orcid.org/0000-0001-5983-9095}{\includegraphics[scale=0.06]{orcid.pdf}}\hspace{1mm}Susmit Jha}\\
    Computer Science Laboratory\\
    SRI International\\
    \texttt{susmit.jha@sri.com} \\
\And
    {Rickard Ewetz} \\
    Electrical and Computer Engineering \\
    University of Central Florida\\
    \texttt{rickard.ewetz@ucf.edu} \\
    \And
{\href{https://orcid.org/0000-0001-6757-105X}{\includegraphics[scale=0.06]{orcid.pdf}}\hspace{1mm}Alvaro Velasquez} \\
    Department of Computer Science\\
    University of Colorado Boulder\\
    \texttt{alvaro.velasquez@colorado.edu} \\
}
\else
\usepackage{authblk}

\newbox{\orcid}\sbox{\orcid}{\includegraphics[scale=0.06]{orcid.pdf}} 
\author[1]{%
	\href{https://orcid.org/0000-0000-0000-0000}{\usebox{\orcid}\hspace{1mm}David S.~Hippocampus\thanks{\texttt{hippo@cs.cranberry-lemon.edu}}}%
}
\author[1,2]{%
	\href{https://orcid.org/0000-0000-0000-0000}{\usebox{\orcid}\hspace{1mm}Elias D.~Striatum\thanks{\texttt{stariate@ee.mount-sheikh.edu}}}%
}
\affil[1]{Department of Computer Science, Cranberry-Lemon University, Pittsburgh, PA 15213}
\affil[2]{Department of Electrical Engineering, Mount-Sheikh University, Santa Narimana, Levand}
\fi

\renewcommand{\shorttitle}{Neural Networks for Robust and Explainable Analysis of Electromagnetic Emissions}

\hypersetup{
pdftitle={Neural Networks for Robust and Explainable Analysis of Electromagnetic Emissions},
pdfsubject={cs.AI, EM},
pdfauthor={Sumit Kumar Jha},
pdfkeywords={Robustness, Interpretability, Explainable Artificial Intelligence, Electromagnetic Signals, Unintended Radiated Emissions, Neural SDEs},
}

\newcommand{\inputx}{\mathbf{x}}
\newcommand{\modelF}{\mathcal{F}}
\newcommand{\attr}{\mathcal{A}}

\npdecimalsign{.}
\nprounddigits{2}

\begin{document}

\maketitle

\begin{abstract}
We present a comprehensive evaluation of the robustness and explainability of ResNet-like models in the context of Unintended Radiated Emission (URE) classification and suggest a new approach leveraging Neural Stochastic Differential Equations (SDEs) to address identified limitations. We provide an empirical demonstration of the fragility of ResNet-like models to Gaussian noise perturbations, where the model performance deteriorates sharply and its F1-score drops to near insignificance at 0.008 with a Gaussian noise of only 0.5 standard deviation. We also highlight a concerning discrepancy where the explanations provided by ResNet-like models do not reflect the inherent periodicity in the input data, a crucial attribute in URE detection from stable devices.
In response to these findings, we propose a novel application of Neural SDEs to build models for URE classification that are not only robust to noise but also provide more meaningful and intuitive explanations. Neural SDE models maintain a high F1-score of 0.93 even when exposed to Gaussian noise with a standard deviation of 0.5, demonstrating superior resilience to ResNet models.  Neural SDE models successfully recover the time-invariant or periodic horizontal bands from the input data, a feature that was conspicuously missing in the explanations generated by ResNet-like models. This advancement presents a small but significant step in the development of robust and interpretable models for real-world URE applications where data is inherently noisy and assurance arguments demand interpretable machine learning predictions. 
\end{abstract}


\section{Introduction}

The unintended radiated emissions from electronic devices can provide a plethora of information to observers about the type of the electronic equipment as well as its current operating condition. Such emissions can be used for activities ranging from non-intrusive load monitoring to side-channel leakage of otherwise secure information. These unintended radiated emissions (UREs) from electronic devices occur due to non-ideal filters, manufacturing variations, and other design constraints, including but not limited to signal modulation, frequency mixing, and high-frequency clocking of signals in digital circuits.

The task of mapping detected UREs to specific devices or specific operating conditions can be thought of as a classification problem in machine learning. Efforts have been placed towards understanding machine learning classification tasks for unintended electronic emissions. In particular, the Oak Ridge National Laboratory has created the Flaming Moe data set~\cite{flamingMoe_osti} of real-world unintended electronic emissions to allow for the design of new URE detection and analysis algorithms. The dataset has been obtained by studying 18 devices and observing two 10-minute segments of voltage data captured at 2 million samples per second. The URE  was detected and measured using high-quality radio equipment to ensure maximum accuracy. Together with the dataset, the team released the   dimensionally aligned signal projection  algorithm as a new approach for creating low-dimensional features for URE classification applications~\cite{vann2017dimensionally}. 

The Flaming Moe data set has also served as a robust benchmark for learning neural network models for URE classification~\cite{grimes2020explanation}. Surprisingly, it has been argued that an out-of-the-box residual neural network model is eminently capable of classifying the URE signals into 18 classes. In fact, a small residual neural network model can achieve perfect accuracy on a held-out fragment on the Flaming Moe data set. This may cause an observer to conclude that further research on neural network-based analysis of the Flaming Moe data set is unnecessary. In this paper, we show that such a conclusion is not true, and the learned residual neural network model begins to fail miserably even in the presence of a small amount of noise in the observed data.

Earlier work on analyzing the Flaming Moe URE data set has realized the importance of communicating the reason for the classification or an explanation from the black-box neural network model to the end human user. In particular, local interpretable model-agnostic explainability has been used to explain  residual neural networks designed for this purpose. 
However, these explanations, as shown in Fig. 4 of~\cite{grimes2020explanation}, do not obey the inductive bias of the data set that the URE signal observed in the short-term Fourier transform is often periodic or time-invariant, and any robust attribution that covers all explanations should uncover this fact. In the case of images, an explanation occurring as a horizontal band in our images uncovers this inductive bias inherent in the data set. We show that modern explanation methods like integrated gradients~\cite{sundararajan2017axiomatic} with smoothgrad~\cite{smilkov2017smoothgrad} uncover this inductive bias in our data set when applied to models based on neural stochastic differential equations.

In summary, the primary contributions of this paper are:

\begin{enumerate}
  \item An empirical demonstration of a lack of robustness to noise for ResNet-like models in the context of URE detection: Our investigation has unearthed certain limitations in the existing residual neural network models. They have been found to be quite fragile, readily falling victim to the addition of Gaussian noise. When perturbed with Gaussian noise of a standard deviation of 0.25, the F1 score drops to a mere 0.41. Increasing the standard deviation to 0.5 results in an almost negligible F1 score of 0.01. See Table~\ref{table:resnet_robustness}.
  \item An empirical demonstration of the lack of periodicity for explanations of short-term Fourier transform images in ResNet-like models: Another concern arises from the data's inherent inductive bias that creates periodic behavior or horizontal bands in the input short-term Fourier transform image. These bands are surprisingly absent in the explanations provided by the ResNet-like models, which are instead interspersed with positive and negative attribution input features in both horizontal and vertical dimensions. See Fig.~\ref{fig:exp_resnet}.
  \item A novel application of models based on neural SDEs in building robust and explainable URE detection models: We present a more robust and explainable framework in the form of Neural Stochastic Differential Equations (SDEs). When compared to the ResNet models, the Neural SDE models are remarkably resilient against Gaussian noise perturbations. For instance, when exposed to Gaussian noise with a standard deviation of 0.5, these models retain an F1 score of 0.93, as compared to the near-zero score of ResNet models. Furthermore, the explanations generated by the Neural SDE models recover the inherent inductive bias in the input, clearly displaying time-invariant horizontal bands. See Table~\ref{table:robust2} and Fig.~\ref{fig:attr_robust}.
\end{enumerate}

In essence, our work is centered around leveraging the power of neural SDEs to create models for URE classification that are more robust to Gaussian noise, explainable, and aligned with the innate properties of the input data.  This innovative approach offers a promising path for the development of sophisticated URE detection algorithms by showing that the current generation of \emph{robust} residual neural network models do not achieve a perfect 100\% accuracy on the Flaming Moe data set, thereby highlighting that this data set remains valuable for building robust explainable neural network models in the future.

\section{Related Work}
\label{sec:bg}

\subsection{Data Collection and Dataset}

\begin{wraptable}{R}{4.5cm}
\small
\centering
\begin{tabular}{|c|}
\hline
\textbf{Devices} \\
\hline
Corelco Phone \\
CyberPower UPS \\
Dell Monitor \\
Dell Optiplex \\
Dell XPS \\
Fluorescent Lights \\
LG Phone \\
Linksys Router \\
Odroid XU4 \\
Polycom VoIP \\
Raspberry Pi \\
Roku 2 XS \\
USRP E310 \\
ViewSonic Monitor \\
Vizio Blu-ray \\
VTech V.Smile \\
Wii U \\
Xbox One \\
\hline
\end{tabular}
\caption{List of Devices}
\label{table:devices}
\vspace{-1cm}
\end{wraptable} To assess the design of machine learning and related classification algorithms, Unintended Radiated Emission (URE) was collected from 18 commercially available electronic devices, commonly found in an office environment. These devices are enumerated in Table I. This Flaming Moe dataset~\cite{flamingMoe_osti}, generated by Oak Ridge National Laboratory in 2016, serves as an idealized URE dataset for the development of URE detection and classification models.

Data collection was organized into four 10-minute segments with the device being operations only in alternate 10-minute windows. Each segment was further split into 1200 files representing 1 second of data for each device. A one minute delay was imposed prior to device capture in order to allow the device to boot and achieve stable performance. This steady behavior creates an inductive bias in our experiments that should be uncovered by any sound and robust explanation approach.

The collection of URE signals for the Flaming Moe data set took place within a Radio Frequency shielded enclosure, employing a USRP N210 collection platform that featured a temperature-compensated crystal oscillator. This oscillator is deemed suitable for low-cost and routine industrial applications due to its modest frequency accuracy of 2.5 ppm, which is adequate for the relatively low-frequency signals observed in our applications.

In order to measure the URE signals, a high impedance differential voltage probe was strategically positioned between the ground and neutral conductors powering the RF chamber. The Ettus Research LFRX analog-to-digital processing board was employed to record the signals. The data collection system employed a twin T-notch filter, which effectively eliminated the 60 Hz component of the signal, ensuring cleaner and more precise data acquisition.

\subsection{ResNet Models for URE}
Recent work~\cite{grimes2020explanation} has argued that a small residual network is capable of achieving a perfect test accuracy on the Flaming Moe data set. We were able to reproduce this rather unusual result in our own experiments. However, we found that inserting a Gaussian noise with a standard
deviation of 0.5 results in an almost negligible F1 score
of 0.01. Hence, the off-the-shelf residual neural network model is very fragile and may not be suitable for real-world data analysis where such non-adversarial noise may be inevitable.

\subsection{Attribution Methods}

Several state-of-the-art attribution methods have been developed over the last decade with increasingly higher degrees of success. However, to the best of our knowledge, we are the first to bring these more contemporary attribution methods to the analysis of neural networks analyzing unintended radiation emission (URE) from devices using the Flaming Moe data set.

Salient methods, like the Layer-wise Relevance Propagation~\cite{bach2015pixel}, decomposes the contribution of each neuron in a network to the final prediction, providing a detailed ``relevance" map. Another technique, known as Shapley Additive Explanations~\cite{lundberg2017unified}, maps each input feature to its numerical importance for a given prediction of the neural network. This approach leverages game theory concepts, attributing the impact of each feature on the neural network response in a way that ensures fairness and consistency.

Integrated Gradients~\cite{sundararajan2017axiomatic}, works by connecting the response of a deep neural network to the features in its input through the concept of path integrals. This is an axiomatic method that provides a simple and intuitive way of understanding the feature attributions. Grad-CAM~\cite{selvaraju2017grad}, employs the gradients or derivatives of a target class from a given convolutional layer. Usually, the gradients from the final layer are used to construct a rough localization map that highlights those features in the input that lead to a given prediction.

More recent research directions are seeking to improve upon these methods, aiming to make them more robust and consistent, to handle complex scenarios with higher reliability. For instance, SmoothGrad~\cite{smilkov2017smoothgrad} and Stochastic Differential Equations~\cite{jha2021smoother} have led to more robust attributions with smaller sensitivity scores for other image data, such as those from ImageNet~\cite{deng2009imagenet}.

\section{Approach}
\label{sec:approach}

Our approach first employs short-term Fourier transform to transform time-series data into visual images. Since Flaming Moe data set has two continuous recordings of 600 seconds with 2 million samples per second, we obtain 2.4 billion samples per device for analysis. We create short-term Fourier transforms using 1 million samples each; thereby, creating 2,400 images per device and obtaining a data set of 43,200 images. For our analysis, we create both deterministic and stochastic variants of residual neural networks, and analyze the robustness and explainability of the model using currently popular attribution methods.

\subsection{Neural Stochastic Differential Equations}

Building upon  advances in modeling neural networks as dynamical systems~\cite{chang2017multi}, our work exploits recent  neural stochastic differential equations (SDEs)~\cite{liu2020does,hodgkinson2020stochastic} extensions of these models to encompass stochastic behavior. In this section, we briefly recall recent results~\cite{liu2020does,jha2021smoother} related to our work. Put succinctly, ResNets can be interpreted as discretizations of neural ordinary differential equations (ODEs) and stochastic variants of residual networks can serve as approximations of neural stochastic differential equations (SDEs). Both the inference and the training processes in ResNets can be represented using dynamical systems~\cite{chen2015learning,chang2017multi,sonoda2017double,chen2018neural,lu2018beyond}. A fundamental ResNet unit, with a residual $R(X(i), W(i))$, can be defined as follows:
\begin{align}
    X(i+1) = X(i) + R(X(i), W(i)) 
\end{align}

\noindent where $X(i)$ denotes the input to the $i^{th}$  block and $X(i+1)$ signifies the block's output that is then fed into the succeeding unit. Here, $W(i)$ indicates the learned weights in the respective residual neural network block. Specifically, in this notation, $X(0)$ denotes the network input $\inputx$ and the network output $\modelF$ is denoted as $X(T)$.
Upon applying suitable limits, we can formulate the evolution of the residual neural network as an ordinary differential equation:

\begin{align}
\frac{dX(t)}{dt} = G(X(t), W(t))
\end{align}

\noindent Here, $G(X(t), W(t)) = \lim_{\delta t \to 0} \frac{R(X(t), W(t))}{\delta t}$ and $X(0)$ is the neural network input.

To model a stochastic variant of the ResNet, a noise term $N(i)$ is added to the right-hand side of the earlier equation.
The  dynamical system for such residual neural networks with a noise component can be represented as an SDE:
\begin{align}
    dX(t) = G(X(t), W(t)) \; dt + \sigma(X(t), t) \;dB(t)
\end{align}
\noindent Here, the noise is depicted as a Brownian motion term $B(t)$, scaled by a suitable diffusion coefficient $\sigma(X(t), t)$.

\begin{figure*}[htb]
\centering
\newcommand{\sizemultiplier}{0.3} %

\begin{subfigure}{\sizemultiplier\textwidth}
  \centering
  \includegraphics[width=\linewidth]{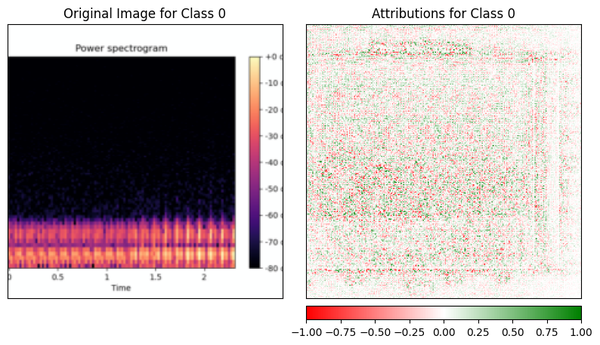}
\end{subfigure}%
~~~~
\begin{subfigure}{\sizemultiplier\textwidth}
  \centering
  \includegraphics[width=\linewidth]{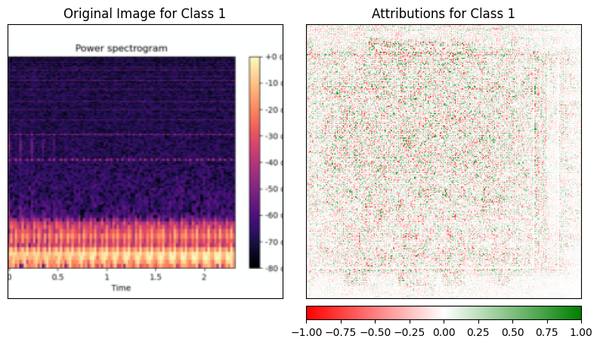}
\end{subfigure}%
~~~~
\begin{subfigure}{\sizemultiplier\textwidth}
  \centering
  \includegraphics[width=\linewidth]{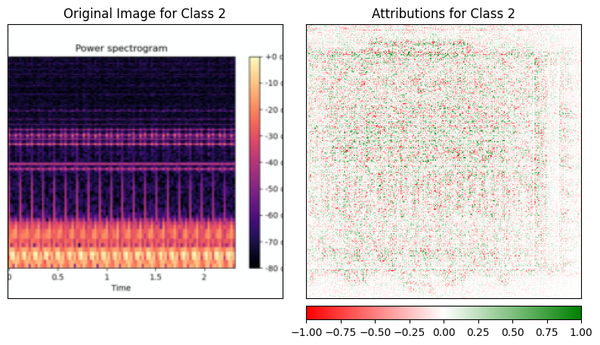}
\end{subfigure}

\begin{subfigure}{\sizemultiplier\textwidth}
  \centering
  \includegraphics[width=\linewidth]{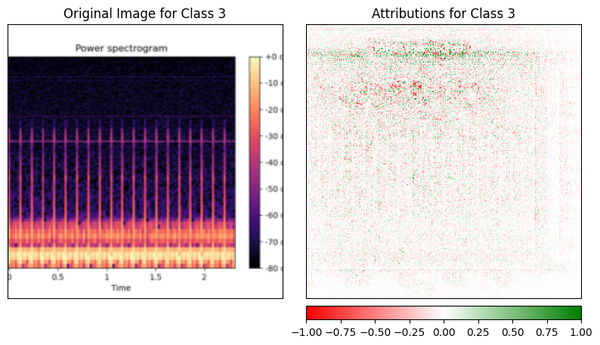}
\end{subfigure}%
~~~~
\begin{subfigure}{\sizemultiplier\textwidth}
  \centering
  \includegraphics[width=\linewidth]{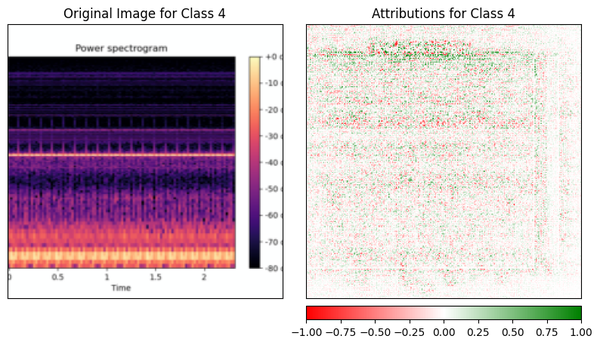}
\end{subfigure}%
~~~~
\begin{subfigure}{\sizemultiplier\textwidth}
  \centering
  \includegraphics[width=\linewidth]{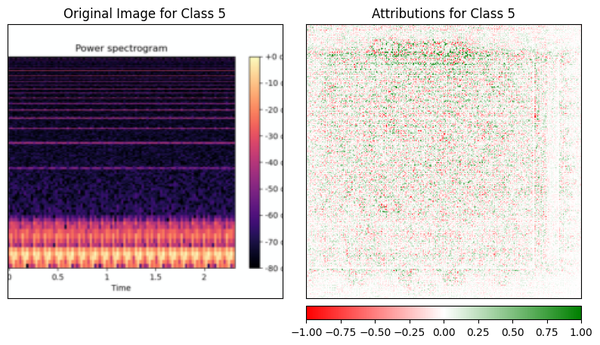}
\end{subfigure}

\begin{subfigure}{\sizemultiplier\textwidth}
  \centering
  \includegraphics[width=\linewidth]{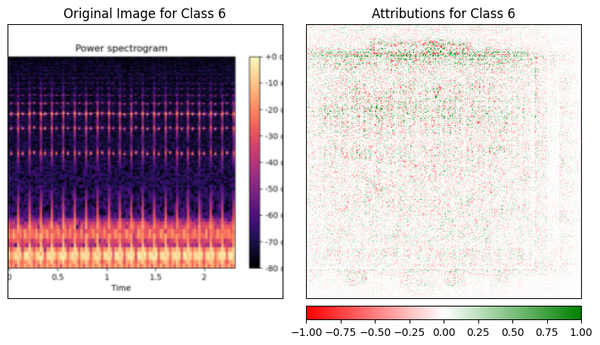}
\end{subfigure}%
~~~~
\begin{subfigure}{\sizemultiplier\textwidth}
  \centering
  \includegraphics[width=\linewidth]{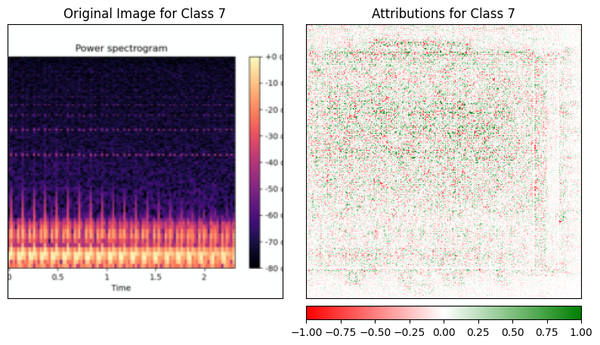}
\end{subfigure}%
~~~~
\begin{subfigure}{\sizemultiplier\textwidth}
  \centering
  \includegraphics[width=\linewidth]{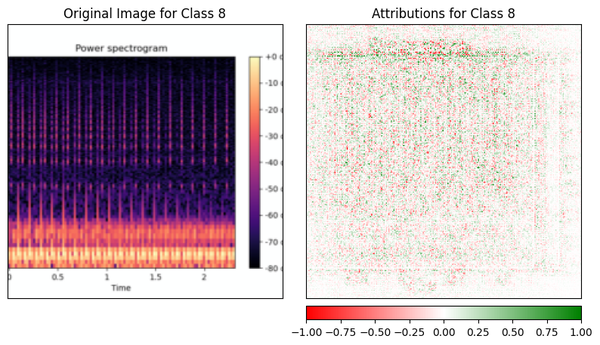}
\end{subfigure}
~~~~
\begin{subfigure}{\sizemultiplier\textwidth}
  \centering
  \includegraphics[width=\linewidth]{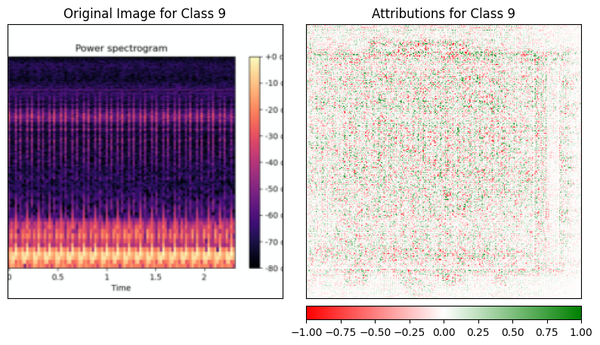}
\end{subfigure}%
~~~~
\begin{subfigure}{\sizemultiplier\textwidth}
  \centering
  \includegraphics[width=\linewidth]{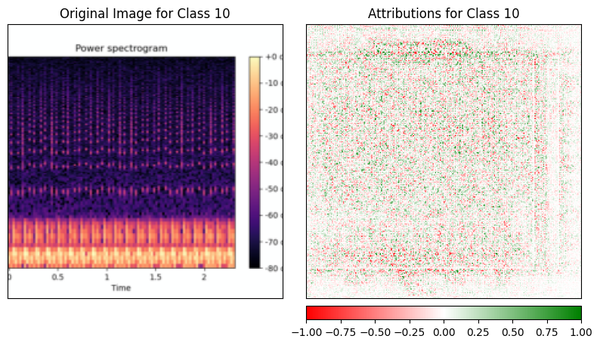}
\end{subfigure}%
~~~~
\begin{subfigure}{\sizemultiplier\textwidth}
  \centering
  \includegraphics[width=\linewidth]{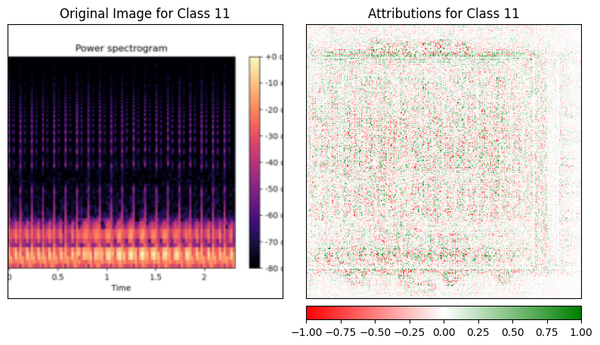}
\end{subfigure}

\begin{subfigure}{\sizemultiplier\textwidth}
  \centering
  \includegraphics[width=\linewidth]{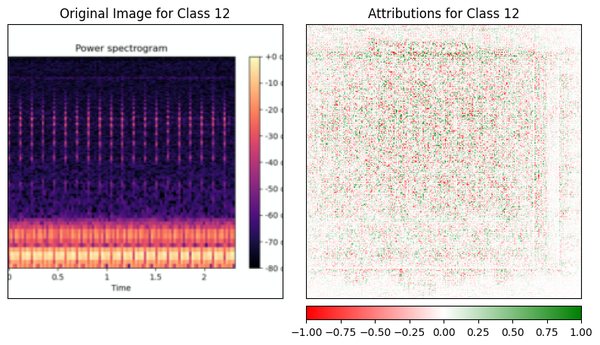}
\end{subfigure}%
~~~~
\begin{subfigure}{\sizemultiplier\textwidth}
  \centering
  \includegraphics[width=\linewidth]{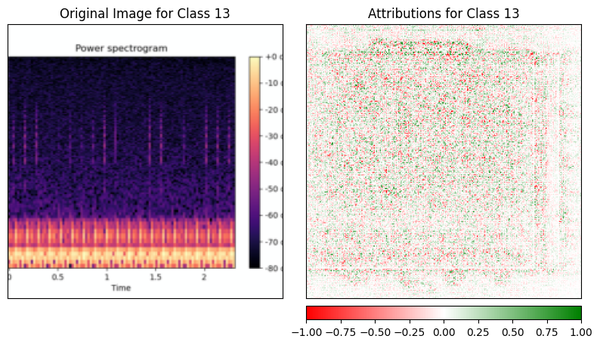}
\end{subfigure}%
~~~~
\begin{subfigure}{\sizemultiplier\textwidth}
  \centering
  \includegraphics[width=\linewidth]{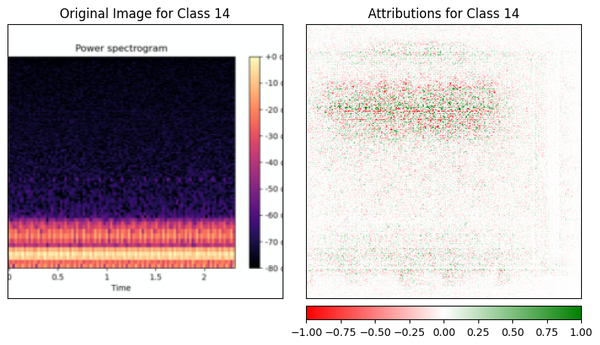}
\end{subfigure}

\begin{subfigure}{\sizemultiplier\textwidth}
  \centering
  \includegraphics[width=\linewidth]{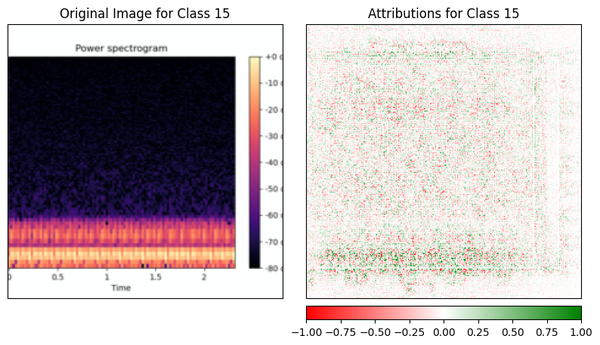}
\end{subfigure}%
~~~~
\begin{subfigure}{\sizemultiplier\textwidth}
  \centering
  \includegraphics[width=\linewidth]{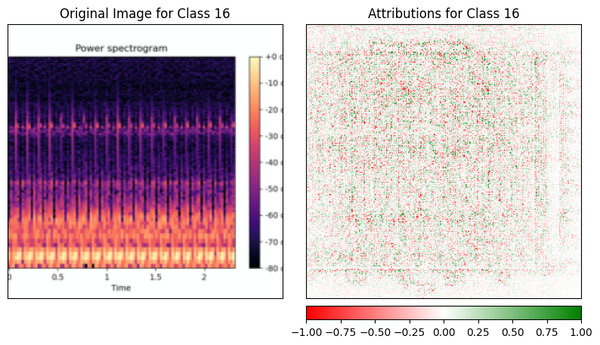}
\end{subfigure}%
~~~~
\begin{subfigure}{\sizemultiplier\textwidth}
  \centering
  \includegraphics[width=\linewidth]{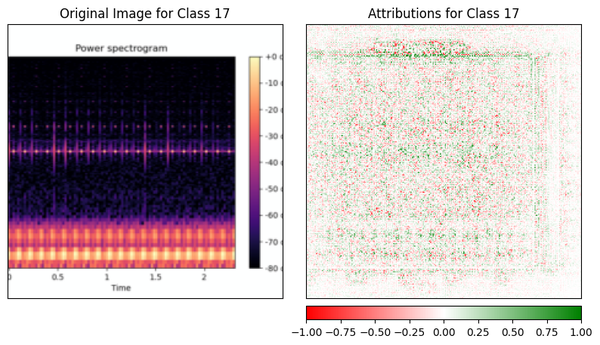}
\end{subfigure}
\caption{ResNet-50 models with perfect accuracy produce attributions that do not support the inductive bias of the data. Horizontal repeated patterns in the data are not identified in the explanations using horizontal bands. Positive (green) and negative (red) attributions are interspersed indicating poor explainability in this context.}
\label{fig:exp_resnet}
\end{figure*}

\subsection{Robust Axiomatic Attributions}

Modern axiomatic methods of attributions, including integrated gradients and their variants, satisfy several fundamental axioms that are not known to be satisfied by methods such as LIME. Integrated Gradients (IG)~\cite{sundararajan2017axiomatic} is an attribution method widely used for feature importance analysis in deep neural networks. The attribution of an input feature in DNNs is typically computed with reference to a baseline input, denoted as $\inputx^b$. 
This baseline could be, for instance, a Gaussian noise image in image-based tasks, or it could be a randomly generated set of inputs. 

Without exhaustively listing all axioms, we highlight a few important axioms to illustrate the importance of integrated gradients and related axiomatic attributions:
\begin{enumerate}
    \item {Linearity:} If a model $F$ is a linear combination of two models, say $F = aF_1 + bF_2$, then the attributions for $F$ should be a linear combination of the attributions for $F_1$ and $F_2$.

    \item {Symmetry-Preserving:} If two variables are symmetric in a model, they should receive the same attribution.

    \item {Implementation Invariance:} If there are two functionally equivalent networks, that is, networks that for the same input give the same output, the attributions should be the same for the equivalent networks.

    \item {Completeness:} The sum of all the feature attributions $\attr_j(\inputx)$ of the model should be equal to the difference between the output of the model $F$ at the input and the baseline. Formally, if $F(\inputx) - F(\inputx^b) = \sum_{j} \attr_j(\inputx)$, then the completeness property is satisfied.  This completeness property is also satisfied by other Shapley value-based methods such as DeepShap~\cite{lundberg2017unified}. 

\end{enumerate}

\begin{table*}[htb]
\centering
\begin{tabular}{cn{1}{2}n{1}{2}n{1}{2}n{1}{2}n{1}{2}n{1}{2}n{1}{2}n{1}{2}n{1}{2}}
\toprule
{Class} & \multicolumn{3}{n{1}{2}}{0.1 Standard Deviation} & \multicolumn{3}{n{1}{2}}{0.25 Standard Deviation} & \multicolumn{3}{n{1}{2}}{0.5 Standard Deviation}\\
\cmidrule(lr){2-4} \cmidrule(lr){5-7} \cmidrule(lr){8-10}
 & Precision & Recall & F1-Score & Precision & Recall & F1-Score & Precision & Recall & F1-Score\\
\midrule
0 & 0.998503 & 0.923823 & 0.959712 & 0.000000 & 0.000000 & 0.000000 & 0.000000 & 0.000000 & 0.000000\\
1 & 0.906133 & 1.000000 & 0.950755 & 0.756178 & 0.633978 & 0.689707 & 0.000000 & 0.000000 & 0.000000\\
2 & 0.998667 & 1.000000 & 0.999333 & 0.997183 & 0.945260 & 0.970528 & 0.000000 & 0.000000 & 0.000000\\
3 & 1.000000 & 0.930459 & 0.963977 & 0.000000 & 0.000000 & 0.000000 & 0.099656 & 0.040334 & 0.057426\\
4 & 0.982783 & 1.000000 & 0.991317 & 0.654369 & 0.983942 & 0.786006 & 0.000000 & 0.000000 & 0.000000\\
5 & 1.000000 & 1.000000 & 1.000000 & 0.139596 & 0.981432 & 0.244426 & 0.000000 & 0.000000 & 0.000000\\
6 & 1.000000 & 1.000000 & 1.000000 & 0.947137 & 0.947137 & 0.947137 & 0.000000 & 0.000000 & 0.000000\\
7 & 0.618729 & 1.000000 & 0.764463 & 0.000000 & 0.000000 & 0.000000 & 0.000000 & 0.000000 & 0.000000\\
8 & 0.987603 & 0.998607 & 0.993075 & 0.357993 & 0.963788 & 0.522067 & 0.000000 & 0.000000 & 0.000000\\
9 & 1.000000 & 1.000000 & 1.000000 & 0.970522 & 0.597765 & 0.739844 & 0.000000 & 0.000000 & 0.000000\\
10 & 0.960474 & 0.990489 & 0.975251 & 0.888889 & 0.021739 & 0.042440 & 0.000000 & 0.000000 & 0.000000\\
11 & 0.991632 & 0.956931 & 0.973973 & 0.547248 & 0.709287 & 0.617819 & 0.000000 & 0.000000 & 0.000000\\
12 & 1.000000 & 1.000000 & 1.000000 & 0.000000 & 0.000000 & 0.000000 & 0.000000 & 0.000000 & 0.000000\\
13 & 1.000000 & 0.581498 & 0.735376 & 0.000000 & 0.000000 & 0.000000 & 0.000000 & 0.000000 & 0.000000\\
14 & 1.000000 & 0.736702 & 0.848392 & 0.000000 & 0.000000 & 0.000000 & 0.000000 & 0.000000 & 0.000000\\
15 & 0.928571 & 0.975469 & 0.951443 & 0.000000 & 0.000000 & 0.000000 & 0.000000 & 0.000000 & 0.000000\\
16 & 1.000000 & 1.000000 & 1.000000 & 1.000000 & 0.677279 & 0.807593 & 0.000000 & 0.000000 & 0.000000\\
17 & 0.996032 & 1.000000 & 0.998012 & 0.925743 & 0.993360 & 0.958360 & 0.043605 & 0.733068 & 0.082314\\
\midrule
Average & 0.964952 & 0.949665 & 0.950282 & 0.454714 & 0.469720 & 0.406996 & 0.007959 & 0.042967 & 0.007763\\
\bottomrule
\end{tabular}
\caption{ResNet model for URE data is fragile to even Gaussian noise. Per-class and average precision, recall, and F1 scores for different standard deviations. The average is reported over all classes.}
\label{table:resnet_robustness}
\end{table*}

While integrated gradients and neural stochastic differential equations have been used to create robust attributions of ImageNet and similar images in the wild, we believe that we are the first to study the use of neural SDEs and integrated gradients on the explainable classification of unintended radiation emissions (UREs) from devices.

\begin{figure*}[t!]
\centering
\newcommand{\sizemultiplier}{0.32} %

\begin{subfigure}{\sizemultiplier\textwidth}
  \centering
  \includegraphics[width=\linewidth]{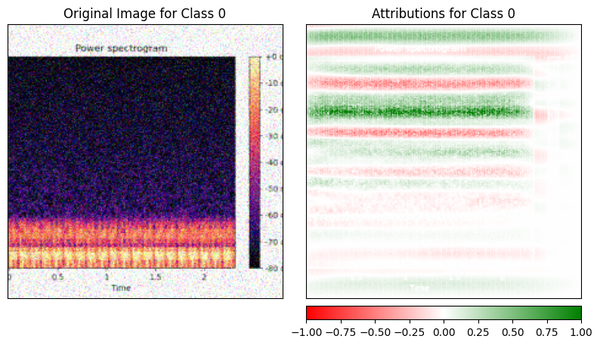}
\end{subfigure}%
~~~~
\begin{subfigure}{\sizemultiplier\textwidth}
  \centering
  \includegraphics[width=\linewidth]{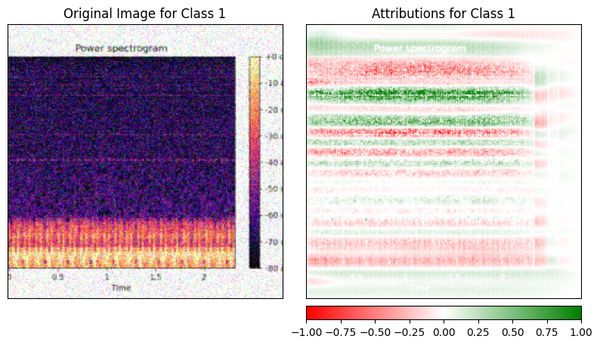}
\end{subfigure}%
~~~~
\begin{subfigure}{\sizemultiplier\textwidth}
  \centering
  \includegraphics[width=\linewidth]{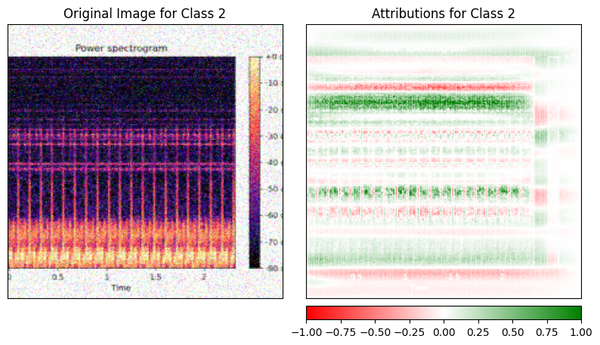}
\end{subfigure}

\begin{subfigure}{\sizemultiplier\textwidth}
  \centering
  \includegraphics[width=\linewidth]{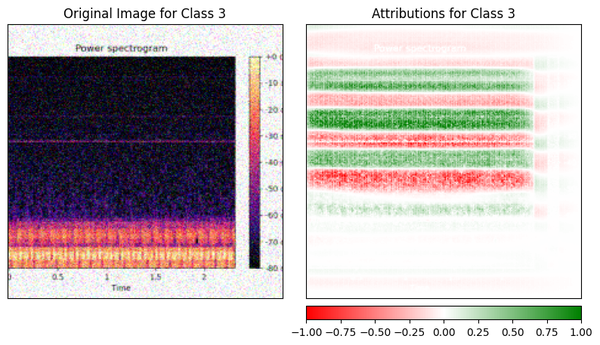}
\end{subfigure}%
~~~~
\begin{subfigure}{\sizemultiplier\textwidth}
  \centering
  \includegraphics[width=\linewidth]{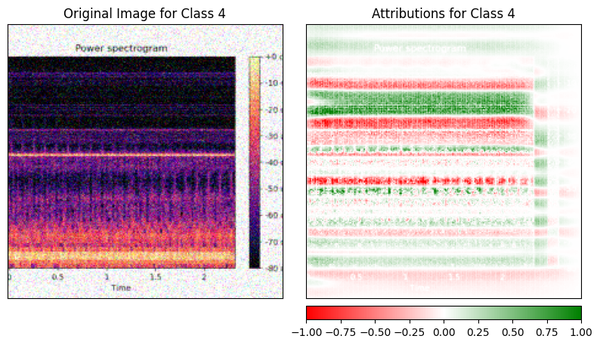}
\end{subfigure}%
~~~~
\begin{subfigure}{\sizemultiplier\textwidth}
  \centering
  \includegraphics[width=\linewidth]{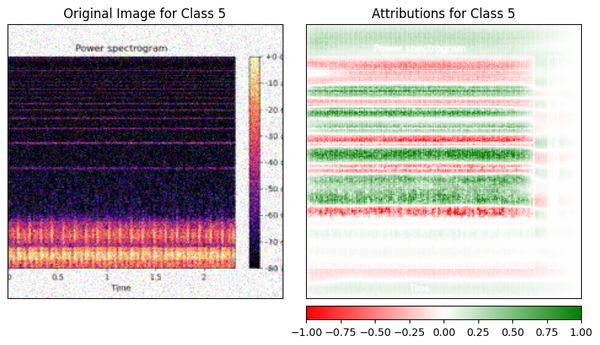}
\end{subfigure}

\begin{subfigure}{\sizemultiplier\textwidth}
  \centering
  \includegraphics[width=\linewidth]{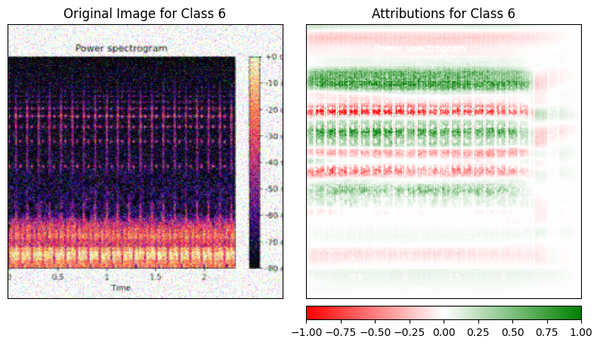}
\end{subfigure}%
~~~~
\begin{subfigure}{\sizemultiplier\textwidth}
  \centering
  \includegraphics[width=\linewidth]{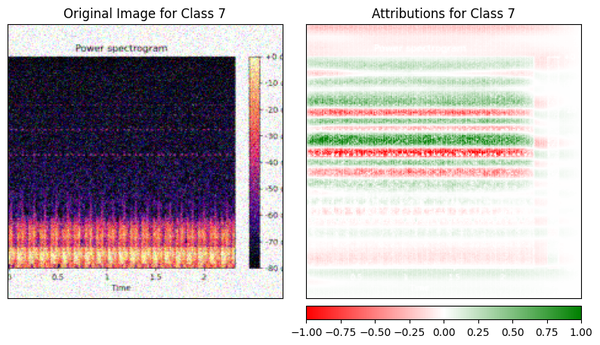}
\end{subfigure}%
~~~~
\begin{subfigure}{\sizemultiplier\textwidth}
  \centering
  \includegraphics[width=\linewidth]{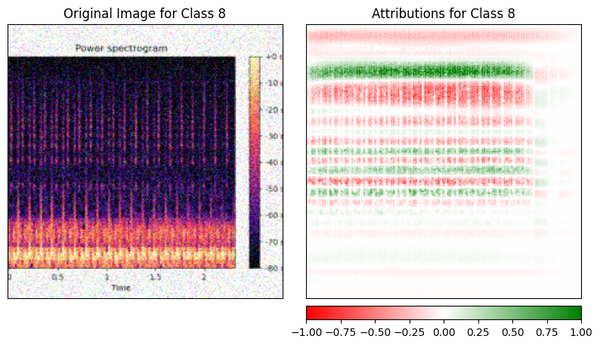}
\end{subfigure}
~~~~
\begin{subfigure}{\sizemultiplier\textwidth}
  \centering
  \includegraphics[width=\linewidth]{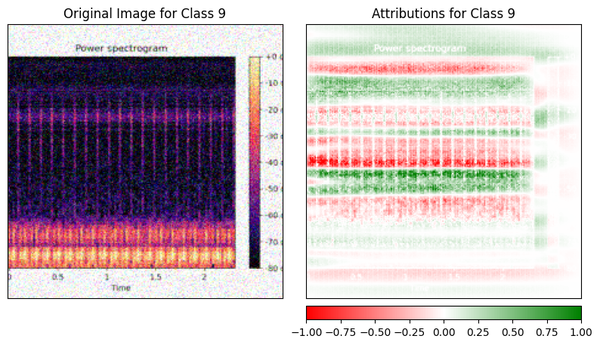}
\end{subfigure}%
~~~~
\begin{subfigure}{\sizemultiplier\textwidth}
  \centering
  \includegraphics[width=\linewidth]{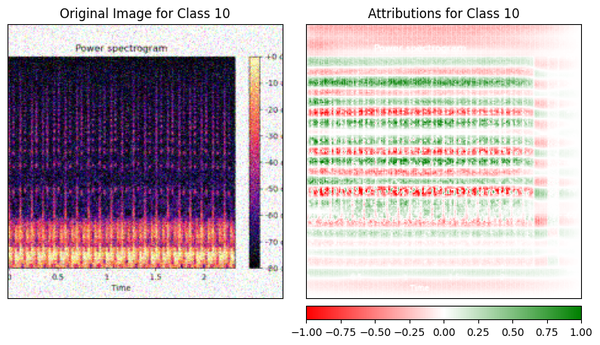}
\end{subfigure}%
~~~~
\begin{subfigure}{\sizemultiplier\textwidth}
  \centering
  \includegraphics[width=\linewidth]{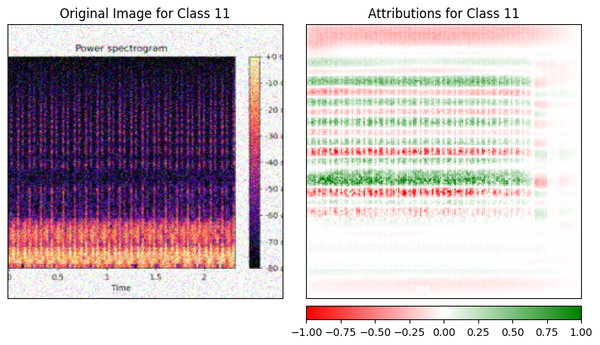}
\end{subfigure}

\begin{subfigure}{\sizemultiplier\textwidth}
  \centering
  \includegraphics[width=\linewidth]{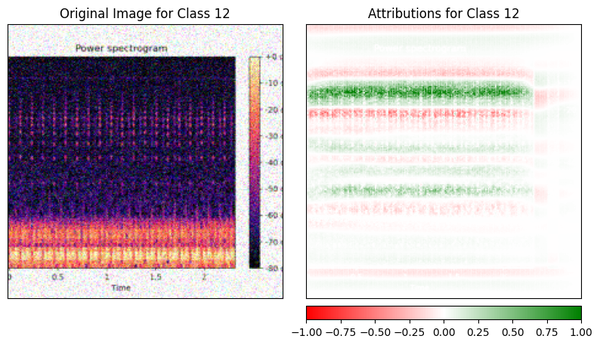}
\end{subfigure}%
~~~~
\begin{subfigure}{\sizemultiplier\textwidth}
  \centering
  \includegraphics[width=\linewidth]{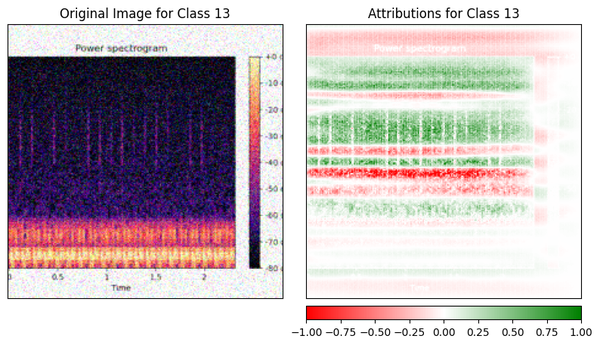}
\end{subfigure}%
~~~~
\begin{subfigure}{\sizemultiplier\textwidth}
  \centering
  \includegraphics[width=\linewidth]{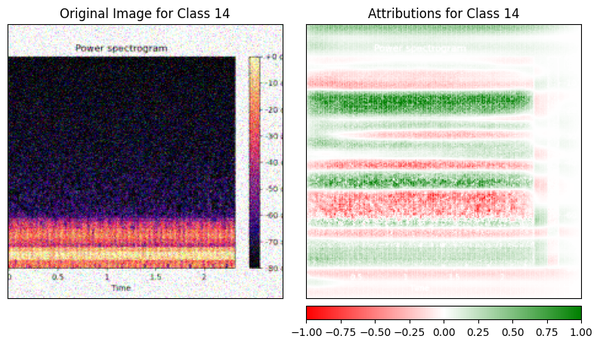}
\end{subfigure}

\begin{subfigure}{\sizemultiplier\textwidth}
  \centering
  \includegraphics[width=\linewidth]{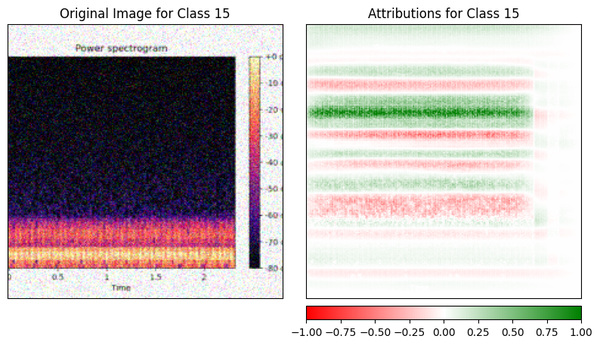}
\end{subfigure}%
~~~~
\begin{subfigure}{\sizemultiplier\textwidth}
  \centering
  \includegraphics[width=\linewidth]{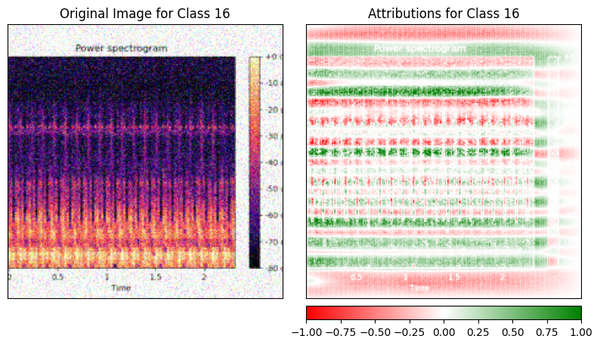}
\end{subfigure}%
~~~~
\begin{subfigure}{\sizemultiplier\textwidth}
  \centering
  \includegraphics[width=\linewidth]{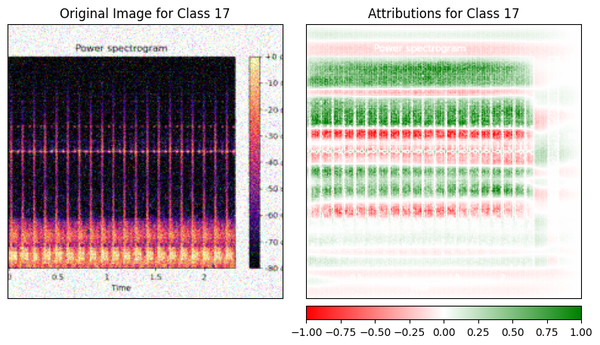}
\end{subfigure}
\caption{Stochastic models  produce attributions that uncover the shape or inductive bias of the input. The positive and the negative attributions occur in horizontal patches that conform to the fact that the data was obtained from a device under stable operation. }
\label{fig:attr_robust}
\end{figure*}

\section{Fragility and Poor Explainability of ResNet}

\subsection{Residual Neural Network Models}

We first employed an image classification model based on the ResNet-50 architecture that was trained on 70\% of the available data and tested on the other held-out 30\% of the data set. The model is designed to classify images into one of 18 different classes. Before being fed into the model, the images are first resized to 224x224 pixels and are then normalized; the model is trained using a batch size of $32$ and the Adam optimizer with a learning rate of $10^{-5}$. The parameters of the model are updated based on the Cross-Entropy loss between the model's predictions and the actual labels.

The training process continues for 10 epochs with the loss going down from $2.7$ to $0.05$. 
The evaluation is performed on a separate test set that constitutes 30\% of the original dataset. The evaluation metrics include precision, recall, and F1-score, and are computed for each class separately, as well as an average over all classes.
The reported results suggest that the model achieves near-perfect performance, with precision, recall, and F1-score of 1.00 for each of the 18 classes.

\begin{table*}[htb]
\centering
\begin{tabular}{cn{1}{2}n{1}{2}n{1}{2}n{1}{2}n{1}{2}n{1}{2}n{1}{2}n{1}{2}n{1}{2}}
\toprule
{Class} & \multicolumn{3}{n{1}{2}}{0.1 Standard Deviation} & \multicolumn{3}{n{1}{2}}{0.25 Standard Deviation} & \multicolumn{3}{n{1}{2}}{0.5 Standard Deviation}\\
\cmidrule(lr){2-4} \cmidrule(lr){5-7} \cmidrule(lr){8-10}
 & Precision & Recall & F1-Score & Precision & Recall & F1-Score & Precision & Recall & F1-Score\\
\midrule
0  & 0.742054 & 0.856135 & 0.795023 & 0.731092 & 0.858956 & 0.789883 & 0.738239 & 0.863188 & 0.795839\\
1  & 0.985392 & 0.998654 & 0.991979 & 0.984064 & 0.997308 & 0.990642 & 0.984085 & 0.998654 & 0.991316\\
2  & 1.000000 & 1.000000 & 1.000000 & 1.000000 & 1.000000 & 1.000000 & 1.000000 & 1.000000 & 1.000000\\
3  & 0.997175 & 0.990182 & 0.993666 & 0.994382 & 0.992987 & 0.993684 & 0.995763 & 0.988780 & 0.992259\\
4  & 1.000000 & 1.000000 & 1.000000 & 1.000000 & 1.000000 & 1.000000 & 1.000000 & 1.000000 & 1.000000\\
5  & 1.000000 & 1.000000 & 1.000000 & 1.000000 & 1.000000 & 1.000000 & 1.000000 & 1.000000 & 1.000000\\
6  & 1.000000 & 0.995828 & 0.997909 & 0.997211 & 0.994437 & 0.995822 & 0.997218 & 0.997218 & 0.997218\\
7  & 0.954733 & 0.929239 & 0.941813 & 0.957650 & 0.935915 & 0.946658 & 0.956885 & 0.918558 & 0.937330\\
8  & 0.978134 & 0.951773 & 0.964774 & 0.972934 & 0.968794 & 0.970860 & 0.977077 & 0.967376 & 0.972202\\
9  & 0.998621 & 1.000000 & 0.999310 & 1.000000 & 1.000000 & 1.000000 & 0.998621 & 1.000000 & 0.999310\\
10 & 0.894172 & 0.814246 & 0.852339 & 0.869369 & 0.808659 & 0.837916 & 0.898911 & 0.807263 & 0.850625\\
11 & 0.774752 & 0.882934 & 0.825313 & 0.771465 & 0.861777 & 0.814124 & 0.779597 & 0.873061 & 0.823686\\
12 & 0.991892 & 1.000000 & 0.995929 & 0.993197 & 0.994550 & 0.993873 & 0.995918 & 0.997275 & 0.996596\\
13 & 0.869754 & 0.803476 & 0.835302 & 0.867733 & 0.798128 & 0.831476 & 0.870504 & 0.808824 & 0.838531\\
14 & 0.852321 & 0.832418 & 0.842252 & 0.874631 & 0.814560 & 0.843528 & 0.855508 & 0.821429 & 0.838122\\
15 & 0.796325 & 0.750361 & 0.772660 & 0.800912 & 0.760462 & 0.780163 & 0.796131 & 0.772006 & 0.783883\\
16 & 1.000000 & 1.000000 & 1.000000 & 1.000000 & 1.000000 & 1.000000 & 1.000000 & 1.000000 & 1.000000\\
17 & 1.000000 & 1.000000 & 1.000000 & 1.000000 & 1.000000 & 1.000000 & 1.000000 & 1.000000 & 1.000000\\
\midrule
Average & 0.935296 & 0.933625 & 0.933793 & 0.934147 & 0.932585 & 0.932702 & 0.935803 & 0.934091 & 0.934273\\
\bottomrule
\end{tabular}
\caption{Stochastic ResNet model is robust to Gaussian noise. Per-class and average precision, recall, and F1 scores for different standard deviations are shown here. The average is reported over all classes.}
\label{table:robust2}
\end{table*}

\subsection{Robustness Analysis}

We investigated the robustness of the learned URE ResNet-50 model under various noise conditions. We synthetically introduced Gaussian noise to the input data with different standard deviations (0.1, 0.25, 0.5), mimicking potential real-world scenarios where data can be distorted due to noise. We evaluated the model's performance across all classes under each noise level, recording the precision, recall, and F1 scores for each class and noise level.

The resulting data, presented in Table~\ref{table:resnet_robustness}, provides an overview of the model's fragility. As the standard deviation of the Gaussian noise increased to 0.25 and 0.5, the performance of the model deteriorated considerably, as evidenced from the decrease in average precision from 0.45 for a standard deviation of 0.25 to 0.01 for a standard deviation of 0.5.

\subsection{Explanation using Integrated Gradients}

We seek to gain an in-depth understanding of the decision-making process of the residual neural network model with perfect accuracy by visualizing the significant features contributing to each prediction. We employed Integrated Gradients~\cite{sundararajan2017axiomatic}, a popular interpretability technique, to identify these important features. We further used a Noise Tunnel~\cite{smilkov2017smoothgrad} with Integrated Gradients to generate smoother attributions and reduce variability in the attributions.
For each of the 18 classes under consideration, we selected one sample that was correctly classified by the model and computed the attributions for the input image.
The computed attributions were then normalized and visualized in the form of heatmaps. The visualization presented two images: the original image and the corresponding heatmap. See Fig.~\ref{fig:exp_resnet}.
We observe that the attributions obtained from the standard residual neural network model do not conform to the inductive bias in the data set that contains nearly periodic or stable signals, and hence should contain horizontal patches in its explanations. In fact, the explanations contain positive and negative attributions next to each other, and have very poor human interpretability.

\section{Results from Stochastic Neural Networks}

\subsection{Stochastic ResNet Model}
\begin{table}[htbp]
\centering
\begin{tabular}{ccccc}
\hline
Class & Precision & Recall & F1-score  \\ 
\hline
0  & 0.75 & 0.86 & 0.80  \\
1  & 0.99 & 1.00 & 0.99  \\
2  & 1.00 & 1.00 & 1.00   \\
3  & 1.00 & 1.00 & 1.00   \\
4  & 1.00 & 1.00 & 1.00   \\
5  & 1.00 & 1.00 & 1.00   \\
6  & 1.00 & 0.99 & 1.00   \\
7  & 0.96 & 0.93 & 0.94   \\
8  & 0.98 & 0.97 & 0.97  \\
9  & 1.00 & 1.00 & 1.00  \\
10 & 0.90 & 0.83 & 0.86 \\
11 & 0.78 & 0.88 & 0.83   \\
12 & 0.99 & 1.00 & 0.99  \\
13 & 0.86 & 0.81 & 0.83   \\
14 & 0.87 & 0.84 & 0.85  \\
15 & 0.82 & 0.78 & 0.80   \\
16 & 1.00 & 1.00 & 1.00 \\
17 & 1.00 & 1.00 & 1.00  \\
\hline
\multicolumn{2}{r}{Accuracy} & \multicolumn{1}{r}{Macro Avg} & \multicolumn{1}{r}{Weighted Avg} \\
\multicolumn{2}{r}{0.94} & \multicolumn{1}{r}{0.94} & \multicolumn{1}{r}{0.94} \\
\hline \\
\end{tabular}
\caption{Test performance of the robust stochastic model. The average is reported over all classes.}
\label{tab:model_performance}
\end{table}

The stochastic ResNet model~\cite{jha2021smoother} for URE data was  evaluated on a test set comprising 12,960 instances across 18 different classes. The robust model displayed a strong overall performance, achieving an accuracy of 0.94. Precision, recall, and F1-score for each class were also examined individually. These results indicate that our model is highly capable of discerning the majority of classes with high accuracy, but efforts could be directed at enhancing its performance on the remaining classes for a more uniformly robust model.
\subsection{Explanations from the Stochastic Model}
We created integrated gradients with noise tunnel explanations~\cite{sundararajan2017axiomatic,smilkov2017smoothgrad} for the stochastic model~\cite{jha2021smoother}, as shown in Fig.~\ref{fig:attr_robust}. The attributions are now completely different from the earlier attributions, and we can now see time-invariant or horizontal patches in the attributions that tell us the frequencies that the model used to classify that input. Green horizontal strips denote the presence of frequencies or the absence of frequencies that caused the model to classify the input into that class. Red horizontal strips denote the presence of frequencies or the absence of frequencies that were counteracting against this classification. 

\begin{figure}[htbp]
    \centering
\includegraphics[width=0.99\textwidth]{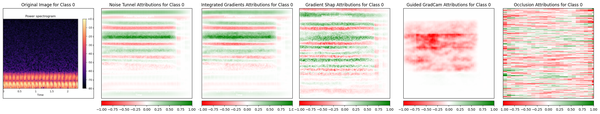}
    \caption{Attributions of the stochastic model for URE data from IG, IG+NT,  GradSHAP, GradCAM and Occlusion methods (left to right).}
    \label{fig:all_attr}
\end{figure}

We  computed IG without noise tunnel, IG with noise tunnel, GradCAM, GradSHAP and occlusion-based attributions for this method to highlight the relative efficacy of the axiomatic integrated gradients approach with noise tunnels compared to other approaches.

\subsection{Robustness Analysis}

The results for the robustness analysis of our robust stochastic model in Table~\ref{table:robust2} can be contrasted with the results obtained for the standard ResNet model, as shown in Table~\ref{table:resnet_robustness}. The standard ResNet model exhibits high performance in terms of precision, recall, and F1-score in the absence of noise. However, this performance drastically degrades as the noise level increases, evident by the average F1-score drops to 0.008, which indicates that the model is almost ineffective in the presence of modest noise levels. In contrast, the robust model demonstrated notable resistance to noise introduction, maintaining an excellent average F1-score of 0.93 even at a noise standard deviation of 0.5. Despite the noisy conditions introduced in the input data, the stochastic model displays a level of performance that was notably higher, emphasizing the utility of our approach for real-world applications where data noise is bound to be prevalent.

\section{Conclusions}

Our investigations have provided hitherto unknown and hopefully valuable insights into the limitations of ResNet-like models when used for Unintended Radiated Emission (URE) detection, particularly their susceptibility to Gaussian noise and the inability of their explanations to capture the inherent inductive bias in the data from a stable device. Our findings underscore the need for more robust and interpretable machine learning models in URE detection.

We have demonstrated that Neural SDEs offer a promising alternative. Not only do stochastic models exhibit remarkable resilience to noise, maintaining high performance even in high-noise scenarios, but they also generate meaningful explanations that capture the inherent inductive biases of the  data. These features make Neural SDE models and their discrete stochastic variants an interesting tool for URE classification.

We identify several opportunities for future research in this direction based on our prior and ongoing work. The Unintended Radiated Emission (URE) classification problem can benefit from the design of hardware solutions with desirable size, weight and power characteristics, such as those based on in-memory computing~\cite{hassen2018free,velasquez2016parallel,pannu2020design} and automated synthesis~\cite{jha2011towards}. Another direction to pursue is to quantify the confidence of the response of the neural network for each instance of URE classifications, using a variety of confidence metrics~\cite{jha2019attribution1,kaur2022idecode}. Neural stochastic differential equations~\cite{jha2022shaping,ijcai2022p853} where the noise has been shaped in conformance with URE data may produce better accuracy and more interpretable results. Analyzing the adversarial robustness~\cite{michel2022survey,jha2019attribution,jha2021protein,jha2018detecting,jha2020model} of robust neural networks trained on URE data remains an open and interesting problem.

\bibliographystyle{alpha}
\bibliography{bib}

\end{document}